\documentclass[journal,twoside,web]{ieeecolor}
\pdfoutput=1
\usepackage{amsmath}
\usepackage{algorithm}
\usepackage{algorithmic}
\usepackage{mathrsfs}
\usepackage{amssymb}
\usepackage{bbm}
\usepackage{array}
\usepackage{eqparbox}

\usepackage{multicol}
\usepackage{multirow}
\usepackage{threeparttable}
\usepackage{makecell}
\usepackage{booktabs}

\usepackage{subfigure}
\usepackage{generic}
\usepackage{cite}
\usepackage{amsmath,amssymb,amsfonts}
\usepackage{algorithmic}
\usepackage{graphicx}
\usepackage{algorithm,algorithmic}
\usepackage{hyperref}
\hypersetup{hidelinks=true}
\usepackage{textcomp}
\def\BibTeX{{\rm B\kern-.05em{\sc i\kern-.025em b}\kern-.08em
    T\kern-.1667em\lower.7ex\hbox{E}\kern-.125emX}}
\begin{document}
\title{Graph Convolutional Network with Connectivity Uncertainty for EEG-based Emotion Recognition}
\author{Hongxiang Gao,~\IEEEmembership{Student Member,~IEEE,}
        Xingyao Wang,~\IEEEmembership{Student Member,~IEEE,}
        Zhenghua Chen,~\IEEEmembership{Senior Member,~IEEE,}
        Min Wu,~\IEEEmembership{Senior Member,~IEEE,}
        Zhipeng Cai,
        Lulu Zhao,
        Jianqing Li,
        Chengyu Liu,~\IEEEmembership{Senior Member,~IEEE}
\thanks{H. Gao, X. Wang, Zhipeng Cai, L. Zhao, J. Li, and C. Liu are with the State Key Laboratory of Digital Medical Engineering, School of Instrument Science and Engineering, Southeast University, Nanjing, 210096. China.}
\thanks{H. Gao, Z. Chen, and M. W are with Institute for Infocomm Research, A*STAR, Singapore, 138632, Singapore.} 
\thanks{J. Li is with the School of Biomedical Engineering and Informatics, Nanjing Medical University, Nanjing, 211166, China.}
\thanks{Z. Chen and C. Liu are the corresponding authors (e-mail: chen0832@e.ntu.edu.sg, chengyu@seu.edu.cn).}}

\maketitle

\begin{abstract}
Automatic emotion recognition based on multichannel Electroencephalography (EEG) holds great potential in advancing human-computer interaction. However, several significant challenges persist in existing research on algorithmic emotion recognition. These challenges include the need for a robust model to effectively learn discriminative node attributes over long paths, the exploration of ambiguous topological information in EEG channels and effective frequency bands, and the mapping between intrinsic data qualities and provided labels.
To address these challenges, this study introduces the distribution-based uncertainty method to represent spatial dependencies and temporal-spectral relativeness in EEG signals based on Graph Convolutional Network (GCN) architecture that adaptively assigns weights to functional aggregate node features, enabling effective long-path capturing while mitigating over-smoothing phenomena. Moreover, the graph mixup technique is employed to enhance latent connected edges and mitigate noisy label issues.
Furthermore, we integrate the uncertainty learning method with deep GCN weights in a one-way learning fashion, termed Connectivity Uncertainty GCN (CU-GCN). We evaluate our approach on two widely used datasets, namely SEED and SEEDIV, for emotion recognition tasks. The experimental results demonstrate the superiority of our methodology over previous methods, yielding positive and significant improvements. Ablation studies confirm the substantial contributions of each component to the overall performance.
\end{abstract}


\section{Introduction}
\label{sec:introduction}
\IEEEPARstart{T}he superior cognition of human intelligence over artificial intelligence is not due to computational capacity but the former's ability to comprehend emotions, a sophisticated function of the brain that mediates self-assessment \cite{cacioppo2000psychophysiology}. This realization has galvanized efforts towards evolving artificial intelligence to discern emotions rather than mere binary judgments. Automated emotion recognition, thus, holds immense promise across varied domains like medical treatment, fatigue monitoring, and human-computer interaction (HCI) systems.

Human emotions, intricate and multi-dimensional, are conveyed through gestures, facial expressions, physiological signals, and more. Among these, physiological signals, particularly those from the brain, have emerged as potent indicators of emotional changes due to their inherent veracity \cite{cacioppo2000psychophysiology}. Electroencephalogram (EEG), along with other physiological measures, has demonstrated to be highly reflective of an individual's emotional states \cite{ullsperger2014neural, naseer2015fnirs}. To unlock this potential, multi-channel emotion EEG induction research is actively pursued, leading to the creation of open-access databases \cite{zheng2015investigating, zheng2018emotionmeter, zhang2018spatial, zhang2020emotion}.


\begin{figure}
    \centerline{\includegraphics[width=3.5in]{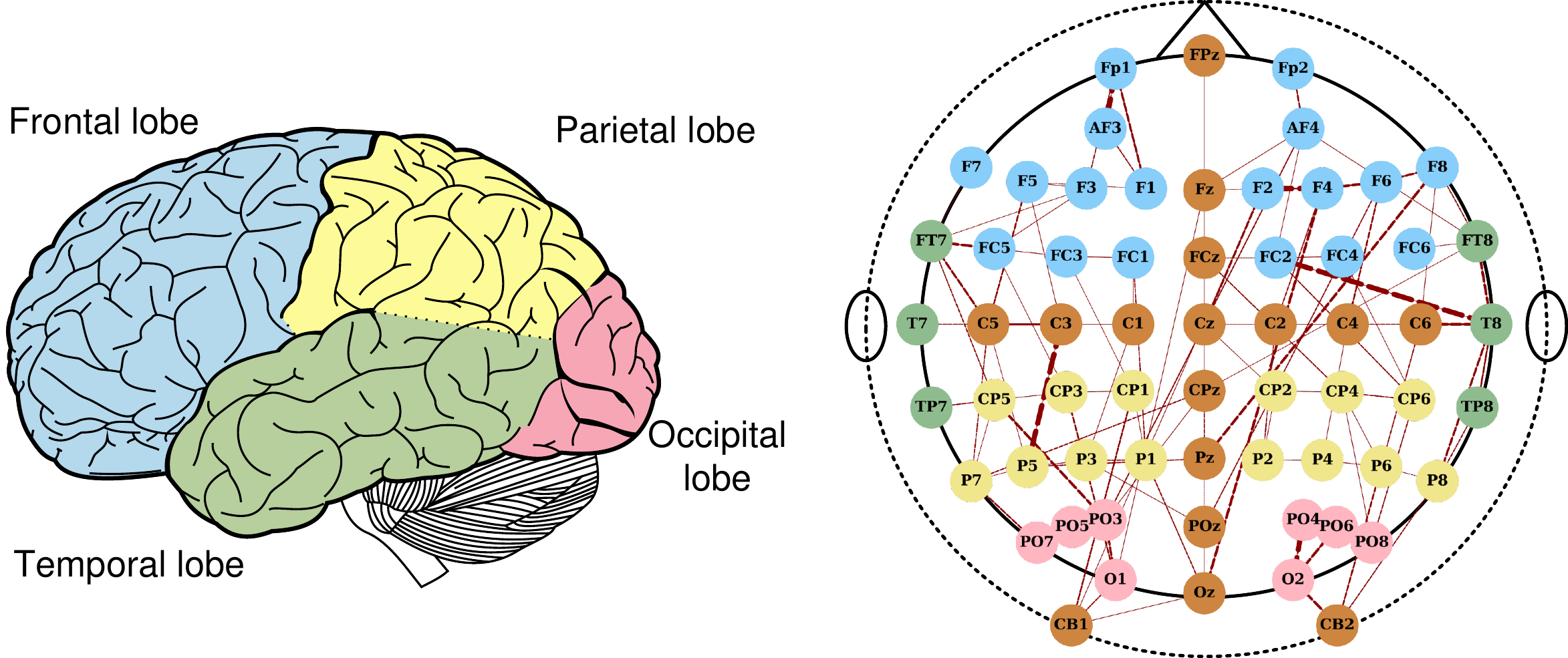}}
    \caption{The illustration of the four brain lobes (left) and the placement of 62 EEG electrodes following the 10-20 systems (right). The correspondences between the four lobes and 62 electrodes are indicated in color. An example of functional connectivity is depicted.}
    \label{fig1}
\end{figure}

Current research delving into brain functional connectivity, as indicated by functional magnetic resonance imaging (fMRI), points towards the potential ability of individuals to maintain attention. EEG-based emotion studies have sought to capture spatial information by transforming the three-dimensional electrode position into a two-dimensional matrix, thereby enabling the use of two-dimensional convolution \cite{jia2020sst}. While conventional models such as CNN and RNN have been employed, their precision has been questioned \cite{britton2006neural}. However, the advent of graph neural networks (GNNs) has allowed treating each EEG channel as a node and connections as edges, leading to significant advancements in EEG-based emotion classification \cite{song2018eeg, zhong2020eeg}.

Despite their popularity, GNNs have been limited by the vanilla graph convolution network (GCN) approach, suffering from over-smoothing beyond a few layers \cite{li2018deeper}. Numerous methods to address this have emerged, some of which include dropout \cite{srivastava2014dropout}, drop-edge \cite{rong2019dropedge}, and residual-based approaches \cite{chen2020revisiting}. While these have alleviated over-smoothing, they often involve computationally expensive operations.

Brain connectivity characterizing emotions requires a nuanced approach. Simple construction of the adjacency matrix using distance coefficients or correlation indices have shown subpar performance in emotion recognition \cite{zhong2020eeg, demir2021eeg, tang2021self}. Emotion processing in the brain is an intricate activity, asynchronous across multiple regions \cite{ledoux2012rethinking}. Positive emotions, for instance, are linked with heightened activity in the left prefrontal cortex, while negative emotions involve the right prefrontal cortex \cite{palmiero2017frontal}. This understanding underscores the need for effective brain functional connectivity construction and long-path dependencies. Current research has thus far had little to offer regarding effective brain functional connectivity construction and long-path dependencies. Figure \ref{fig1} represents an example of 62-channel electrode distribution and possible connection among different brain lobes.

\begin{figure}
    \centerline{\includegraphics[width=3.0in]{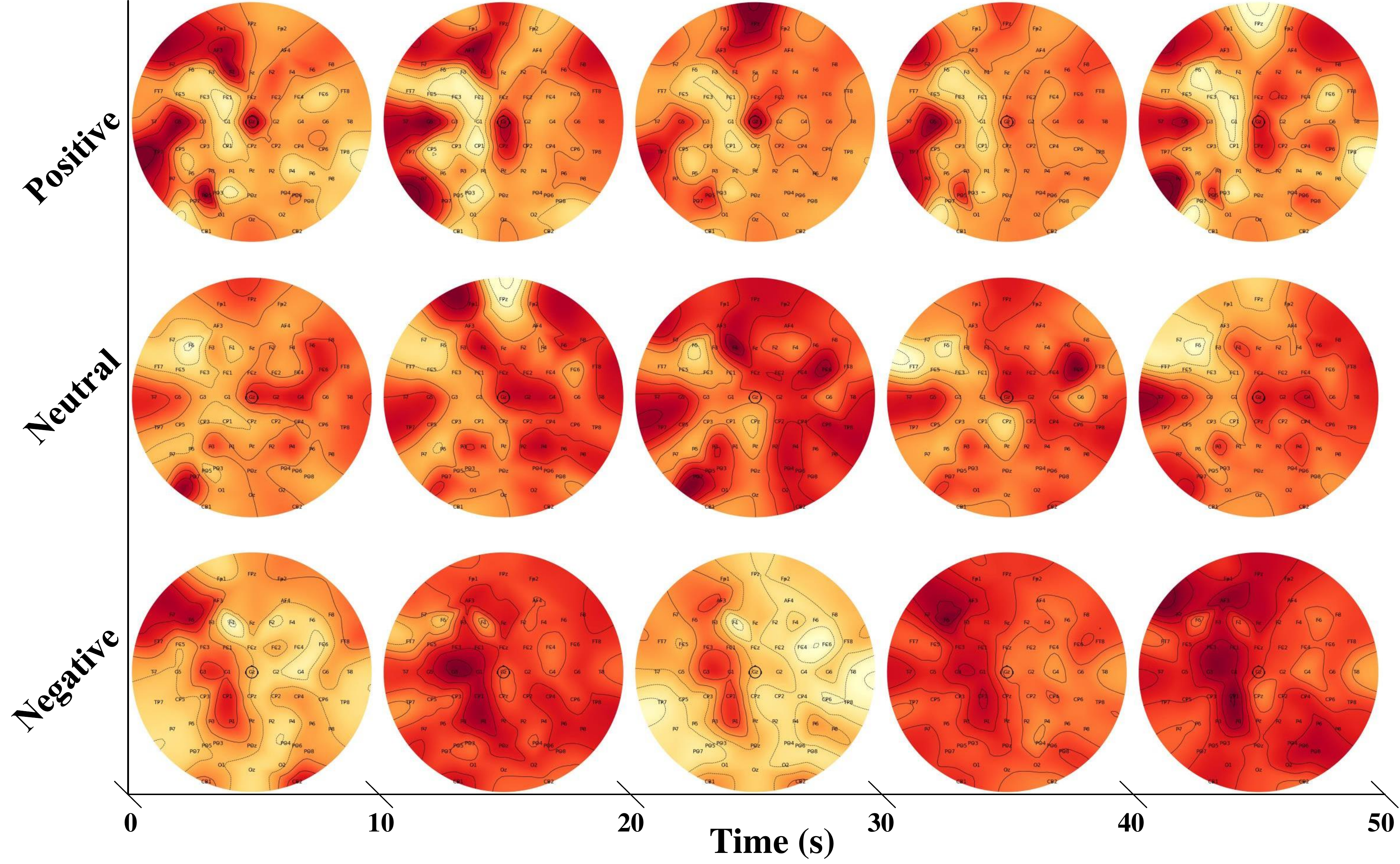}}
    \caption{An Example of Temporal Brain Activation Maps for Three Different Emotions in the Same Individual in SEED.}
    \label{fig2}
\end{figure}

Restricted to the collection difficulties of EEG signals, all available datasets are designed in a lab environment. During trials that involve video-induced emotions, participants often encounter challenges such as insufficient stimulation or extreme variability, resulting in varying emotional states throughout the test. As depicted in Figure \ref{fig2}, even for the same individual, the brain activation patterns for the same emotion over continuous periods of time exhibit noticeable variations, despite maintaining some consistent activation regions. Traditionally, one-second EEG signals have been treated as individual samples, using handcrafted features for emotion categorization \cite{zhong2020eeg, song2018eeg, zheng2018emotionmeter, zheng2015investigating, song2021variational}. However, accurately labeling and identifying emotions based on these short segments pose difficulties. This issue has been overlooked by many researchers, except for Zhong \textit{et al.}, who introduced the concept of soft labels based on the spatial relationship between different emotions on the arousal-valence plane \cite{zhong2020eeg}. Their approach provides a more nuanced understanding of video-induced emotions, overcoming the limitations of traditional single-sample annotations and enhancing comprehension in this domain.

Despite the aforementioned strategies, the efficacy of GCN may be constrained in certain scenarios due to the following factors: 1) Intrinsic limitations stemming from over-smoothing issues, impeding effective long-path information acquisition; 2) Inadequate exploration of the brain's spatial topological structure and the significance of spectral features; and 3) Insufficient induction resulting in data-label discrepancies, thereby constraining the effectiveness of strategies.


In our endeavor to address limitations in EEG emotion recognition, we have introduced an uncertainty-guided deep Graph Convolutional Network (GCN) model. Our model efficiently mitigates the challenges of over-smoothing and noisy labels by leveraging an uncertainty-guided approach for adjacency matrix generation and Bayesian techniques to estimate connectivity probability, which enhances connectivity between local and global brain regions. We have also demonstrated that the learning of connection uncertainty within our GCN parallels the PageRank algorithm, an advancement that enables the elimination of over-smoothing issues and captures a broader range of spectral information. Furthermore, we have employed a data mixing technique to address the issue of noisy labels, thus introducing potentially connected edges for robust representation. The main contributions of our work lie in the following aspects:

\begin{enumerate}
   \item We introduce an uncertainty-driven Graph Convolutional Network (UC-GCN), effectively addressing functional connectivity uncertainties and frequency band obscurities in EEG signals.
    \item We illustrate the role of uncertainty masks in our spectral-enhanced GCN structure, mitigating over-smoothing and augmenting high-frequency information capture.
    \item We innovate with a graph mixing augmentation strategy, enhancing the model's capability to discriminate ambiguous data through potentially connected edges and soft labels.
\end{enumerate}

\section{Related Works}
\subsection{EEG based emotion classification}

EEG signals are a key resource for assessing human emotional states. In the past, feature extraction for emotion categorization was primarily manual, with time domain characteristics such as statistical measures \cite{tang2017eeg}, higher-order correlation methods \cite{lan2016real}, and event-related potentials \cite{olofsson2008affective} being used. From a frequency domain perspective, features like power spectral density were used \cite{lin2010eeg, zheng2015investigating}, along with techniques such as short-time Fourier transform and discrete wavelet transform for time-frequency domain features \cite{liu2017real, sorkhabi2014emotion}. Nonlinear dynamics and chaos-based concepts have also been examined \cite{van1998dynamical}, with nonlinear features such as fractal dimension \cite{li2015assessing} and multifractal detrended fluctuation analysis \cite{paul2015eeg} showing significant correlations with different emotional states.

Despite their value, these features have mostly been focused on single-channel signals, ignoring the relationships between different channels. This is a significant oversight, as research has shown that different parts of the brain can be affected differently by emotional states \cite{jones1992electroencephalogram}. To address this, recent research by Shi et al. \cite{shi2013differential} used differential and rational asymmetry measures to quantify differences in hemispheric brain activity. Tao et al. \cite{tao2020eeg} proposed using an attention-based convolutional recurrent neural network to extract both channel relationships and inherent similarities from EEG data. Li et al. \cite{9105104} also used directed recurrent neural networks on two hemisphere areas to capture both spatial and temporal dependencies simultaneously. Graph neural networks (GNNs) have been employed by researchers to analyze intra-channel connections, with each node representing inter-channel characteristics \cite{song2018eeg, song2021variational, zhong2020eeg}. While promising, these GNNs often use predetermined adjacency matrices that lack a robust clinical or psychological basis.

\subsection{Graph Neural Network in EEG}
Traditional neural networks like CNNs and RNNs face challenges in non-Euclidean spaces, pushing researchers to explore alternatives. Spectral methods using Laplace transformed space for convolution have been introduced but were computationally demanding \cite{estrach2014spectral}. This was improved with Chebyshev polynomials, enhancing graph convolutional networks (GCNs) \cite{defferrard2016convolutional}. Spatially-based methods, like defining fixed neighbor vectors for graph convolution or building adjacency matrices via randomly selected neighbors, have also been proposed \cite{niepert2016learning, hamilton2017inductive}. However, deep GCNs tend to oversmooth, leading to indistinct representations \cite{li2018deeper}. Solutions like custom PageRank matrices have been suggested to combat this issue \cite{gasteiger_predict_2019}.

Regarding EEG data tasks, GNNs have shown promising results in emotion categorization \cite{zhong2020eeg, song2018eeg, song2021variational}, epilepsy identification \cite{li2021spatio}, and seizure analysis \cite{tang2021self}. Most of the research focuses on constructing brain functional connectivity based on thresholds or constructing a graph that combines local and global regions. However, these methods often rely on a predefined graph, which may not fully capture the interconnectedness of different emotional states across individuals. Also, static connection graphs may struggle to represent brain activation patterns accurately in noisy environments. As a solution, we suggest an uncertainty-driven strategy for replacing the adjacency matrix, which integrates weighted root node features to mitigate over-smoothing and improve model performance for all emotions.

\begin{figure*}[!ht]
    \centerline{\includegraphics[width=6.6in]{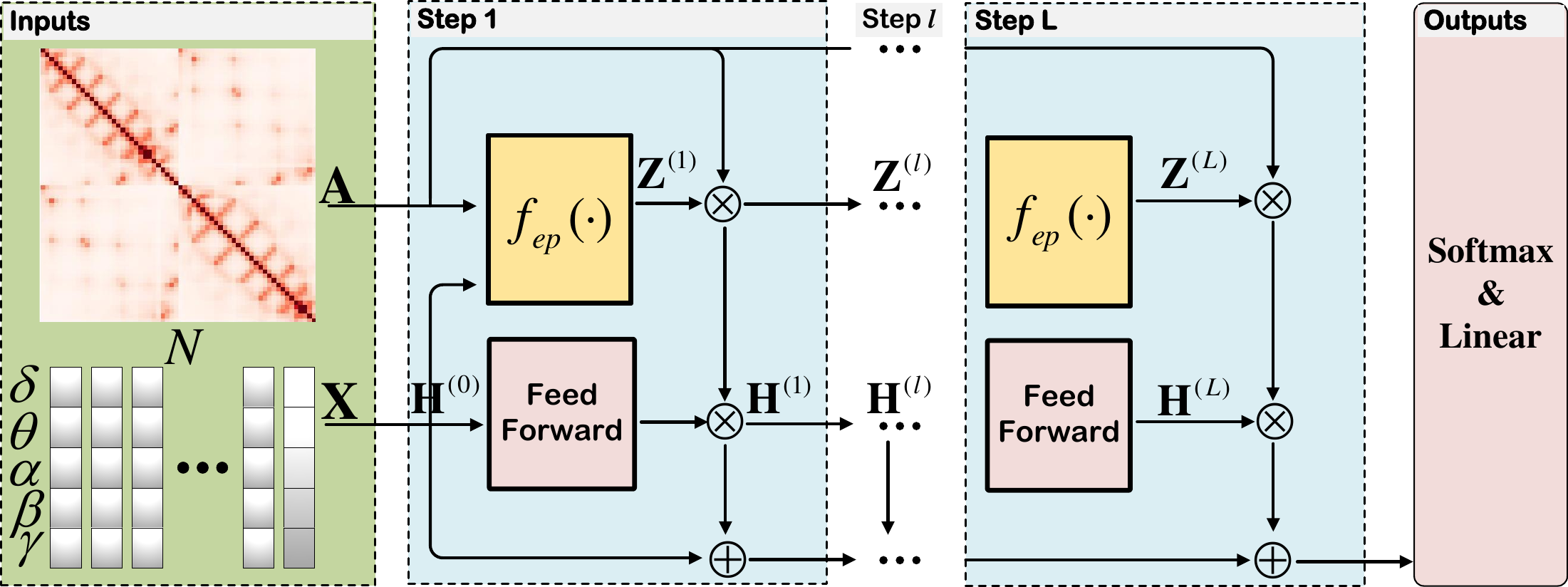}}
    \caption{The framework of our Connectivity Uncertainty GCN. For a given adjacency matrix ${\bf A} \in \mathbb{R}^{N \times N}$ and node features ${\bf X} \in \mathbb{R}^{N \times f_l}$, we generate each layer a binary mask ${\bf Z}^{(l)} \in \mathbb{R}^{N \times N \times f_l} (l=1, 2,..., L-1)$ using an edge predictor $f_{ep}(\cdot)$ and element-wise multiplied with $\bf A$. The transition matrix for each layer ${\bf H}^{(l)}$ was allocated a weighted coefficient $\alpha_l$ to introduce both local and global information to better describe the final output.}
    \label{fig3}
\end{figure*}

\subsection{Uncertainty for graphs}

While uncertainty quantification in CNNs has been extensively studied, it has received limited attention in the context of GNNs \cite{abdar2021review, gao2022novel}. Aleatoric uncertainty, arising from imprecise and noisy measurements, affects the observability of node characteristics, edge connectivity, and edge weights in a graph. Zhang \textit{et al.} \cite{zhang2019bayesian} proposed a Bayesian framework that generates graph structures and parameters using parametric random graphs to handle aleatoric uncertainty. Munikoti \textit{et al.} \cite{munikoti2022general} established a generic Bayesian framework and employed assumed density filtering to measure aleatoric uncertainty. Epistemic uncertainty, on the other hand, arises from the model's limited ability to accurately represent the underlying process. Variational inference \cite{kingma2015variational, song2021variational} and sampling-based approaches \cite{gal2016dropout} are commonly used to estimate posterior density functions of model parameters. Hasan \textit{et al.} \cite{hasanzadeh2020bayesian} presented an adaptive connection sampling-based stochastic regularization algorithm for GNNs, while Munikoti \textit{et al.} \cite{munikoti2022general} and Feng \textit{et al.} \cite{feng2021uag} employed Monte-Carlo dropout during testing to quantify prediction uncertainty. Existing efforts in graph data analysis have primarily focused on feature qualification and model performance, assuming pre-defined graph structures. However, for EEG-related activities, the functional connectivity of the brain remains a challenging mystery. Our study explicitly incorporates uncertainty in graph structures into our paradigm, addressing this crucial aspect.

\section{Methods}
In this section, we then present our key idea of automatically learning the connections between different EEG channels to reconstruct the brain's functional graph for emotional response, as shown in Figure \ref{fig3}.

\subsection{Preliminaries}
\subsubsection{Graph Neural Networks}
We represent an undirected graph $\mathcal{G}=(\mathcal{V}, \mathcal{E})$ with node set $\mathcal{V} = \{v_1, v_2, \ldots, v_N\}$ and edge set $\mathcal{E} \subseteq \mathcal{V} \times \mathcal{V}$. The edges in the graph may be alternately represented using an adjacency matrix ${\bf A} \in \mathbb{R}^{N \times N}$, where ${\bf A}_{ij} = 0$ indicates that node $i$ and $j$ are not connected and $N = |\mathcal{V}|$ is the number of nodes. Let the node attribute matrix be ${\bf X} \in \mathbb{R}^{N \times f}$, where $f$ denotes the input feature dimension and each node may be endowed with a $f$-dimensional node attribute vector ${\bf X}_{i:}$. Let ${\bf D}=diag(d_1, \ldots, d_N)$ be the diagonal degree matrix of ${\bf A}$ with $d_i=\sum_j {\bf A}_{ij}$.

\textbf{Spectral Graph Convolution \cite{defferrard2016convolutional} -} Spectral graph theory \cite{chung1997spectral} study the graph topology property by means of the eigenvalues and eigenvectors of the Laplace matrix of graphs. Laplacian matrix is defined as $\bf L = D-A$ and its normalization matrix is $\bf \widetilde L = I - D^{-1/2}AD^{-1/2}$, which is a semi-definite matrix with eigenvalues as $\bf \Lambda$ and eigenvectors as $\bf U$. Thus, $\bf \widetilde L = U\Lambda U^T$. Similar to the role of CNNs in image processing, spectral convolution on graphs can be understood as a filtering operation in the spatial domain. It is achieved by performing a multiplication between the raw graph signal $\mathbf{X}$ and the filter kernel $\mathbf{G}$ in the spectral domain, utilizing techniques such as Fourier transform \cite{defferrard2016convolutional} or wavelet transform \cite{xu2018graph, zheng2021framelets}. This spectral convolution operation allows us to extract meaningful features from graph-structured data, analogous to the way CNNs extract features from images.
\begin{equation}
    \bf X*G = U\bigg((U^TG) \odot (U^TX)\bigg)=U \hat GU^TX,
\label{Eq1}
\end{equation}
where $\odot$ denotes the element-wise multiplication and ${\bf \hat G} = \text{diag}(\hat g_1, \ldots, \hat g_N)$ denotes a diagonal matrix of spectral filter coefficients. To avoid the high computation overload of matrix decomposition of $\bf \widetilde L$, \cite{defferrard2016convolutional} approximated $\bf \hat G$ with $K$-order Chebyshev polynomials $T_k(\bf \hat{\Lambda})$:
\begin{equation}
    {\bf \hat G_{\Lambda}} \approx \sum_{k=0}^{K}\theta_kT_k({\bf \hat \Lambda}),
    \label{Eq2}
\end{equation}
and the corresponding filter coefficients $\hat{g}(\lambda_j)=\sum_k \theta_k \lambda_j^k$, where ${\bf \hat \Lambda} = \frac{1}{2 \lambda_{max}} {\bf \Lambda} -{\bf I}$ is a scaled Laplacian normalization, $\lambda_{max}$ denotes the largest eigenvalue of $\bf L$ and $\theta \in \mathbb{R}^K$ is a learnable vector of Chebyshev coefficients, $k \in \{0, \dots, K \}$ means the $k$-th term of a $K$-order polynomials. 

\textbf{Graph convolution neural network (GCN) \cite{kipf2016semi}} further approximate the Eq. (\ref{Eq2}) with $K=1, \theta_0=2$, and $\theta_1=-1$, so that the convolution operation in Eq. (\ref{Eq1}) translate into $\bf X \ast G = (I + D^{-1/2}AD^{-1/2})X$. Each GCN layer transformation function $\bf H$ is defined as:
\begin{equation}
    \begin{aligned}
        {\bf H}^{(l+1)} &= \sigma\bigg(\tilde {\bf D}^{-\frac{1}{2}} {\bf \tilde{A}} {\bf \tilde D}^{-\frac{1}{2}}{\bf H}^{(l)}{\bf W}^{(k)}\bigg) \\
            &= \sigma\bigg({\bf \tilde A}_{sym} {\bf H}^{(l)}{\bf W}^{(l)}\bigg),
    \end{aligned}
    \label{Eq3}
\end{equation}
where ${\bf \tilde A}_{sym} = {\bf \tilde D}^{-\frac{1}{2}} {\bf \tilde{A}} {\bf \tilde D}^{-\frac{1}{2}}$ is a normalized adjacency matrix, $\bf \tilde{D}=I+D, \tilde{A} =I + A$ is a renormalization trick introduced in GCN, ${\bf H}^{(0)}={\bf X}$, ${\bf W}^{(l)}$ is the convolutional parameter of each layer to realize the feature transformation, $\sigma(\cdot)$ denotes a nonlinear activation function and $k$ denotes the layer number. The normalized Laplacian ${\bf \tilde L}_{sym} = {\bf I - \tilde D}^{-\frac{1}{2}} {\bf \tilde{A} \tilde D}^{-\frac{1}{2}} = {\bf \tilde U \tilde{\Lambda} \tilde U^T}$ and its corresponding graph filter can be described as $\hat{g}(\hat{\lambda}_i) = {(1-\hat{\lambda}_i)}^L$, where $\hat{\lambda}_i$ are eigenvalues of ${\bf \tilde A}_{sym}$. 

For brevity and without loss of generality, we denote $\bf A$ as the adjacency matrix with a self-loop and $\bf D$ as the degree matrix with a self-loop.

\subsubsection{Over-smoothing in GNNs}
The aggregation progress of neighborhood information can be seen as a graph filtering step over node features with a filter expressed in Eq. (\ref{Eq2}). Obviously, the preserved spectral component in $\bf \hat{G}_{\Lambda}$ is dominated by the eigenvalues of $T_k(\bf \hat{\Lambda})$, i.e. ${\bf \hat{\Lambda}} = \{\hat{\lambda}_1, \ldots, \hat{\lambda}_K \}$ and $\hat{\lambda}_1 >, \ldots, > \hat{\lambda}_K $, which in turn represent the low to high-frequency components. Currently, practical models are usually shallow (number of layers $K$ is 2 - 4), as such a choice offers the best empirical performance which is later proved attributed to the low-pass characteristic. Besides, the transition matrix will finally collapse to an independent matrix of input without a linear interposition in Eq. (\ref{Eq3}), $ \lim\limits_{k \to \infty} {\bf H}^{(k)} = {\bf H}^{(\infty)} {\bf \leftarrow} {\bf \tilde{A}}_{sym}^{\infty}{\bf H}^{(0)}$. And the final decision derived from such an undiscriminating, ${\bf Y}_{pred}= {\rm softmax}(\sigma({\bf H}^{(L)}{\bf W}))$ is merely determined by the $L$-hop node attributes with $L$ denotes the total layers, provided that the graph is irreducible and aperiodic. More specifically, the learned representations will lose discriminative information provided by nodes with different topological and feature characteristics as the model goes deeper.

\subsection{Uncertainty-guided GCN for Emotion Recognition from EEG Signals}

\subsubsection{Emotion-specific function connectivity reconstruction}
In practical applications, each EEG channel is regarded as a node, with node vectors encapsulating the computed features across various frequency bands, specifically, the delta, theta, alpha, beta, and gamma bands.
Our central proposition rests on leveraging the information derived from both the inherent topology of the graph in the spatial domain and the node features in the time-frequency domain to overcome prevalent challenges such as the identification of topological connections and distinct effective frequency band responses. This two-pronged approach allows us to pinpoint crucial information to be assimilated into the propagation of knowledge during the design phase of the GCN.

A crucial step in this procedure is the introduction of an edge predictor (EP), which computes the connection probabilities between any two nodes, paving the way for node feature learning. The edge predictor can be formally expressed as $f_{ep}: \boldsymbol{\rm A, X} \rightarrow \boldsymbol{\rm Z}$, where $\boldsymbol{\rm Z}$ denotes an edge existence mask.
We leverage the mask matrix ${\bf Z}$ to establish deterministic links to high-scoring non-edges and remove poorly scored linked edges. The propagation procedure is then reformulated as the following metric:
\begin{equation}
    \begin{aligned}
        {\bf H}^{(l+1)} &= \sigma \left( {\bf D}^{-\frac{1}{2}} ({\bf A} \odot {\bf Z}^{(l)}) {\bf D}^{-\frac{1}{2}} {\bf H}^{(l)} {\bf W}^{(l)} \right).
    \end{aligned}
\label{Eq5}
\end{equation}
It is easy to observe that implementing node-wise and feature-wise uncertainty quantization leads to different probabilistic interpretations. To provide a comprehensive measure of model uncertainty, we further extend ${\bf Z}^{(l)} = {\bf Z}_{uv}^{(l)} \cdot {\bf Z}_{f}^{(l)}$ as a 3-dimensional matrix, where ${\bf Z}_{uv}^{(l)} \in \mathbb{R}^{N \times N \times 1}$ denotes the connection of any two nodes $u, v$, and ${\bf Z}_{f}^{(l)} \in \mathbb{R}^{N \times f_{l}}$ denotes the mask vector of layer $l$ on features of each node. We can rewrite and reorganize Eq. \ref{Eq5} in a feature-wise view as
\begin{equation}
    \begin{aligned}
        {\bf H}^{(l+1)} &= \left\{{\bf H}_{[:, j]}^{(l+1)} \in \mathbb{R}^{N \times N}\right\}_{j=1}^{f_{l+1}}, \\
    \end{aligned}
\end{equation}
where
\begin{equation}
    {\bf H}_{[:, j]}^{(l+1)} = \sigma\left(\sum_{i=1}^{f_l} {\bf D}^{-\frac{1}{2}} ({\bf A} \odot {\bf Z}^{(l)}[:,i]) {\bf D}^{-\frac{1}{2}} {\bf H}^{(l)}[:, i] {\bf W}^{(l)}[i, j] \right),
\end{equation}
that means we allocate each frequency component with a different topological connection strategy for a better group of useful information. ${\bf Z}^{(l)}[:, i]$ is a binary mask sampled from a Bernoulli distribution with a success rate as $p_l$ for each layer. The success rate $p_l$ is adaptive for different representations, guaranteeing the effectiveness of topology construction.

\subsubsection{Uncertainty Learning With Bayesian Approximation}

To generate an uncertain binary mask ${\bf Z}^{(l)}$, we endeavor to leverage the power of Bayesian inference to train our predictor, using the predictive posterior derived from training data. Intriguingly, we illustrate in the context of this study that the dynamic process of adaptive connection learning can be elegantly transposed from the realm of output feature space to the more abstract parameter space, thereby casting it as a credible Bayesian process encapsulated within the Graph Convolutional Network (GCN) framework.

To further clarify our approach, we revisit Eq. \ref{Eq5} and reinterpret it from a node-wise perspective of a GCN layer, which can be described as:
\begin{equation}
    {\bf H}^{(l+1)} = \left\{{\bf H}_{u}^{(l+1)} \in \mathbb{R}^{N \times f_{l+1}}\right\}_{u=1}^N,
\end{equation}
where
\begin{equation}
    \begin{aligned}
        {\bf H}_{u}^{(l+1)} &= \sigma\left({d_u}^{-\frac{1}{2}} \Bigg(\sum_{v \in \mathcal{N}_u} {\bf A}_u \odot {\bf Z}^{(l)}[v,:]\Bigg) {d_u}^{-\frac{1}{2}} {\bf H}^{(l)}_u {\bf W}^{(l)} \right) \\
        &= \sigma\left({d_u}^{-\frac{1}{2}} \Bigg(\sum_{v \in \mathcal{N}_u} {\bf Z}_{uv}^{(l)} \odot {\bf H}^{(l)}_u {\bf W}^{(l)} \Bigg){d_u}^{-\frac{1}{2}} \right) \\
        &= \sigma\left({d_u}^{-\frac{1}{2}} \Bigg(\sum_{v \in \mathcal{N}_u} {\bf H}^{(l)}_u \text{diag}({\bf Z}_{uv}^{(l)}) {\bf W}^{(l)} \Bigg){d_u}^{-\frac{1}{2}} \right) \\
        &:= \sigma\left({d_u}^{-\frac{1}{2}} \Bigg(\sum_{v \in \mathcal{N}_u} {\bf H}^{(l)}_u {\bf W}^{(l)}_{uv} \Bigg){d_u}^{-\frac{1}{2}} \right).
    \end{aligned}
\end{equation}
In this context, we introduce the notation ${\bf W}^{(l)}_{uv} = \text{diag}({\bf Z}_{uv}^{(l)}) {\bf W}^{(l)}$, thereby facilitating the transference of the learning process for the uncertainty-based edge predictor to the acquisition of a weighted adjacency matrix along the edges $e = (u, v) \in \mathcal{E}$ for each respective layer.

The model parameters that require optimization are denoted as $\boldsymbol{\omega} = \{\boldsymbol{\omega}_l\}_{l=1}^L$, with $\boldsymbol{\omega}_l = \{{\bf W}_e^{(l)}\}_{e=1}^{|\mathcal{E}|}$ serving as the weight set for the $e$-th edge. Here, $|\mathcal{E}|$ represents the number of edges. Initially, each row of ${\bf W}_e^{(l)}$ is assigned a distribution according to $p(\boldsymbol{\omega})$. Moreover, we postulate a vector ${\bf m}_l$ of dimensions $f_l \times f_{l-1}$ for each layer. Consequently, the predictive probability of the deep GNN model can be expressed as:
\begin{equation}
    p(y|{\bf A, X}) = \int_{\boldsymbol{\omega}} p(y|\boldsymbol{\omega}, {\bf A, X}) p(\boldsymbol{\omega} | {\bf A, X})d{\boldsymbol{\omega}}.
    \label{Eq6}
\end{equation}

\subsubsection{Variational Interpretation}
However, the posterior distribution $p(\boldsymbol{\omega}|{\bf A, X})$ proves intractable. In light of this, we adopt a variational interpretation, where a binary mask (each row of $\boldsymbol{\omega}$) can be viewed as an approximate distribution $q_{\theta}(\boldsymbol{\omega})$ that approximates the posterior distribution $p(\boldsymbol{\omega}|{\bf A, X})$. We define $q_\theta(\boldsymbol{\omega})$ as follows:
\begin{equation}
    \begin{aligned}
        {\bf W}_e &= {\bf m}_l \cdot \text{diag}([{\bf z}_{l, e}]_{e=1}^{|\mathcal{E}|}), \\
        {\bf z}_{l, e} &\sim \text{Bernoulli}(p_l) \ \text{for} \quad l = 1,..., L, \ e = 1,...,|\mathcal{E}|.
    \end{aligned}
    \label{Eq7}
\end{equation}
Here, $\theta = \{p_l, {\bf m}_l\}_{l=1}^L$ represents the variational parameters, and ${\bf m}_l$ denotes the mean weight matrices. The variable ${\bf z}_{l, e} = \{0, 1\}^{f_l}$ corresponds to the removal of features, leading to the omission of nodes in layer $l-1$ as connected neighbors within the graph ${\bf A}^{(l)}$, solely when the entire set of features is eliminated.

To optimize the aforementioned model, we introduce the Kullback-Leibler (KL) divergence ${\rm KL}(q_\theta(\boldsymbol{\omega})||(p(\boldsymbol{\omega}))$ in order to minimize the discrepancy between the approximate posterior $q_\theta(\boldsymbol{\omega})$ and the prior distribution $p(\boldsymbol{\omega})$. The discrete quantized Gaussian prior distribution is commonly employed to analytically evaluate this intractable KL divergence. Considering the distribution of the edges, we have $q_\theta(\boldsymbol{\omega})=\prod_{l=1}^L\prod_{e=1}^{|\mathcal{E}|}q_{\theta_l}({\bf W}_e^{(l)})$. By approximating each edge distribution as $q_{\theta_l}({\bf W}_e^{(l)})=p_l \delta({\bf W}_e^{(l)}-0)+(1-p_l)\delta({\bf W}_e^{(l)}-{\bf m}_l)$, we can express the general KL term as:
\begin{equation}
{\rm KL}(q_{\theta_l}({\bf W}_e^{(l)}) || p({\bf W}_e^{(l)})) = \sum_{l=1}^L\sum_{e=1}^{|\mathcal{E}|}{\rm KL}(q_{\theta_l}({\bf W}_e^{(l)}) || p({\bf W}_e^{(l)})),
\label{Eq8}
\end{equation}
Consequently, the edge predictor loss $\mathcal{L}_{ep}$ can be approximated as:
\begin{equation}
\mathcal{L}_{ep} = {\rm KL}(q_{\theta_l}({\bf W}_e^{(l)}) || p({\bf W}_e^{(l)})) \propto \frac{(1-p_l)}{2}{||{\bf m}_l||}^2-\mathcal{H}(p_l),
\label{Eq9}
\end{equation}
where $\mathcal{H}$ denotes the entropy of a Bernoulli random variable with success rate $p_l$, defined as:
\begin{equation}
\mathcal{H}({p}_l)=-p_l{\rm log}p_l-(1-p_l){\rm log}(1-p_l).
\label{Eq10}
\end{equation}

Given a fixed edge connection probability $p_l$, the entropy term remains unaffected by model weights and can be viewed as a constant regularization term during optimization, thereby rendering it negligible. However, the connectivity sampling probability inherently relies on the specific functional structure of each frequency component. The minimization of the KL divergence term equates to maximizing the entropy term with a probability of $1-p_l$, which is optimally achieved when $p_l$ approximates 0.5. It becomes apparent that constant proportion sampling fails to meet the requirements for intricate functional connections and may introduce extraneous information. In order to train $p_l$ during the model training process, we must compute the derivative of the entropy objective, which presents a formidable challenge.

Several commonly used estimators, such as Reinforce \cite{williams1992simple} or pathwise derivative re-parametrization methods \cite{kingma2013auto, kingma2015variational, titsias2014doubly}, are inapplicable due to the discrete nature of the masks. Rather than sampling the random variable from a discrete Bernoulli distribution, we employ an approximation by substituting the discrete distribution with its continuous relaxation. This can be seen as a relaxation of the "max" function in the Gumbel-max trick to a "softmax" function. The transition from a discrete Bernoulli random variable $\bf z$ to a continuous variable $\bf {\tilde z}$ is reparametrized as $\bf {\tilde z} = g(\theta, \epsilon)$, where $\theta$ represents the model parameters and $\epsilon$ is a random variable independent of $\theta$. Assuming that $\epsilon$ follows a uniform distribution, i.e., $\epsilon \sim \text{Unif}(0,1)$, we utilize the Sigmoid function to constrain the continuous value within the interval [0, 1]:
\begin{equation}
{\bf \tilde z} = {\rm Sigmoid}\bigg(\frac{1}{t}\Big({\rm log}\Big({\frac{p}{1-p}}\Big)+{\rm log}\Big({\frac{\epsilon}{1-\epsilon}}\Big)\Big)\bigg),
\label{Eq11}
\end{equation}
where $t$ is a temperature parameter. This distribution assigns most of its mass to the boundaries of the interval 0 and 1. With this concrete relaxation of the adjacency masks, we can now optimize the connection probability.

\label{sec3.3}

\subsection{Alleviating Over-smoothing and Enhance Spectral Filter}
Long path information, known as the high-frequency component among the graph, was demonstrated to have strong discriminative power for graph-level classification. Hence a natural research direction of research regarding GNNs is to investigate how to leverage long paths over graphs without over-smoothing the vertex features. Generalized PageRank (GPR) values \cite{li2019optimizing, chien2020adaptive} enable more accurate characterizations of hidden state distance and similarities and hence lead to improved performance of various graph learning techniques. Given a seed feature ${\bf H}^{(0)}$ and another deeper hidden feature ${\bf H}^{(l)}$ in the graph, the GPR value is deﬁned as $\sum_{l=0}^\infty \alpha_l {\bf H}^{(l)}$, for some GPR weight sequence $\{\alpha_l\}_{l \geq 0} \in {\mathbb{R}}$. 
We rewrite the multi-step propagation with a $L$ truncated polynomial based on Eq \ref{Eq5}:
\begin{equation}
    \begin{aligned}
         {\bf Y}_{pred} &= {\rm softmax}\left(\Big(\sum_{l=0}^{L-1} \alpha_l {\bf H}^{(l+1)}\Big){\bf W}^{(l)} \right) \\
                  &= {\rm softmax}\left(\Big(\sum_{l=0}^{L-1} \alpha_l {\bf D}^{-\frac{1}{2}} ({\bf A} \odot {\bf Z}^{(l)}) {\bf D}^{-\frac{1}{2}}\Big){\bf H}^{(0)}{\bf W}^{(l)} \right), \\
            {\bf H}^{(0)}  &= \bf X,
    \end{aligned}
    \label{Eq15}
\end{equation}
noted that the task of learning GPR score $\alpha$ for graph classification can be reformulated as learning a flexible factor over the adjacency matrix in different propagation steps, thus could be learned with ${\bf Z}^{(l)}$, i.e., ${\tilde{ \bf Z}}^{(l)} = \alpha_l \mathbbm{1}_{N \times N} \odot {\bf Z}^{(l)}$. That means the long-path learning could be realized with the uncertainty learning on each step.
Thus, learning the optimal uncertainty mask ${\bf Z}^{(l)}$ is equivalent to learning the optimal polynomial graph filter. As any graph filter could be approximated with a polynomial graph filter and increasing $L$ allows one to better approximate the underlying feature extractor. The uncertainty-guided GNN is able to leverage long-path topological structure information and with an adaptive learned GPR weight, one can enhance the beneficial middle state while suppressing the harmful.

\textbf{Augmented Spectral Filters}. 
More specifically, ${\bf D}^{-\frac{1}{2}} ({\bf A} \odot \mathbbm{1}_{N \times N} \odot {\bf Z}^{(l)}) {\bf D}^{-\frac{1}{2}} = {{\bf \tilde U} g_{\hat \alpha, L}({\bf \Lambda}) {\bf \tilde U}^T}$ corresponds to a polynomial graph filter of order $L$ over ${\bf H}^{(0)}$. Wherein the $\hat \alpha$ is a multiplier factor derived from ${\bf Z}^{(l)}$ ranged in (0, 1) as proved in Eq. \ref{Eq11}. The corresponding polynomial graph filter equals to $\hat{g}_{\hat \alpha, L}(\hat{\lambda}_i) = {(1-\hat{\alpha}_l \hat{\lambda}_i)}^L$. As proved in \cite{kipf2016semi, wu2019simplifying}, the eigenvalues of renormalized GCN range in [0, 1.5]. When $\hat{\lambda}_i=0$, the amplitude of vanilla GCN is equal to robust GCN, i.e. $g(0) = \hat{g}_{\hat \alpha, L}(0) = 1$. Hence, the robust GCN retains the low-pass filter ability. However, when $\hat{\lambda}_i > 0$, $\hat{g}_{\hat \alpha, L}(\hat{\lambda}_i)=(1-\hat{\alpha}_l\lambda_i)>g(\hat{\lambda}_i)=(1-\hat{\lambda}_i)$, which provides the model more high frequency components. Besides, as $\hat \alpha_l$ is less than 1 and will be smaller with the stacking of the convolution layer, $\hat{g}_{\hat \alpha, K}$ will hardly get a negative value, which is previously proved harmful for the model performance \cite{wu2019simplifying}, thus enhance the robustness of such a model. We refer to the proposed model as Enhanced Graph Convolutional Network (eGCN).




\subsection{Implicit adversarial training strategy}
In section \ref{sec3.3}, we sparsify adjacency ${\bf A}$ with Bernoulli-based uncertainty sampling to get the graph variant adjacency ${\bf A} \odot {\bf Z}^{(l)}$. However, the generated adjacency matrix is able to keep the most class-specific connections but weak in adding global emotion-effective connections, which we argue should be generally existing edges for all emotional states. Besides, existing methods generate samples by segmenting raw data with a non-overlapping 1-s window, ignoring the fact that participants may not always be effectively stimulated when watching an induced material due to self-resistance or insufficient stimulation. Unwanted data are adversarial samples that are harmful to robust model training.

We propose a data-agnostic augmentation method to add latent connected edges and alleviate the mismatching problem (noisy label) by leveraging information from other positive samples of the same mini-batch. In a nutshell, we use mixup \cite{zhang2018mixup} to construct virtual training examples. Given a pair of graph samples $\mathcal{G}_i$, $\mathcal{G}_j$ with the embedding ${\bf X}_{\mathcal{G}_i}, {\bf X}_{\mathcal{G}_j}$, initial adjacency matrix ${\bf A}_{\mathcal{G}_i}, {\bf A}_{\mathcal{G}_j}$ and its corresponding label ${\bf Y}_i, {\bf Y}_j$, we interpolate contextual information by
\begin{equation}
    \begin{aligned}
        \hat{\bf X}_{\mathcal{G}_i\mathcal{G}_j} &= \beta {\bf X}_{\mathcal{G}_i} + (1-\beta){\bf X}_{\mathcal{G}_j}, \\
        \hat{\bf A}_{\mathcal{G}_i\mathcal{G}_j} &= \beta {\bf A}_{\mathcal{G}_i} + (1-\beta){\bf A}_{\mathcal{G}_j}, \\
        \hat{\bf Y}_{ij} &= \beta {\bf Y}_i + (1-\beta){\bf Y}_j,
    \end{aligned}
    \label{Eq14}
\end{equation}
with $\beta \in [0,1]$ controlling the strength of interpolation between feature-target pairs.
We sample the mixup weight $\beta$ from the Beta distribution ${\rm Beta}(\psi, \psi)$ with $\psi$ as a hyper-parameter \cite{zhang2018mixup, wang2021mixup, gupta2004handbook}.  With such an augmentation technique, the converted graph data tend to be close to its intrinsic distribution and labeled with a more convicting soft label. 
Let $\mathbf{h}_{\mathcal{G}}$ denote the graph embedding and the optimization objective of graph classification can be written as: 
\begin{equation}
    \begin{aligned}
        \mathcal{L}_{gc} = \sum \nolimits_{ij} \bigg(\beta {\rm log}p({\bf Y}_i|{\bf h}_{\mathcal{G}_i}, \theta) + (1-\beta){\rm log}p(\hat{{\bf Y}_{ij}}|{\hat {\bf h}}_{\mathcal{G}_{i}\mathcal{G}_j}, \theta)\bigg).
    \end{aligned}
    \label{Eq13}
\end{equation}

Consequently, our emotion recognition method was designed by optimizing both the classification loss $\mathcal{L}_{gc}$ and uncertainty edge predictor loss $\mathcal{L}_{ep}$ in Eq. (\ref{Eq9}) 
\begin{equation}
    \mathcal{L} = \mathcal{L}_{gc} + \mathcal{L}_{ep}.
\end{equation}

\begin{algorithm}[t]
    \renewcommand{\algorithmicrequire}{\textbf{Input:}}
	\renewcommand{\algorithmicensure}{\textbf{Output:}}
    \caption{Connectivity Uncertainty Graph Neural networks}
    \label{alg1}
    \begin{algorithmic}[1]
        \REQUIRE Given a pair of graphs with node attributes and labels $({\bf X}_{\mathcal{G}_i}, {\bf A}_{\mathcal{G}_i}, {\bf Y}_i), ({\bf X}_{\mathcal{G}_j}, {\bf A}_{\mathcal{G}_j}, {\bf Y}_j)$ of a mini-batch, concrete distribution temperature weight $t$, Beta distribution weight $\psi$, number of epochs $T$, batch size $B$.
        \ENSURE Prediction.
        \STATE Initialize model parameters;
        \FOR{$t = 0 \to {T-1}$}
            \STATE Draw two samples $({\bf X}_{\mathcal{G}_i}, {\bf A}_{\mathcal{G}_i}, {\bf Y}_i), ({\bf X}_{\mathcal{G}_j}, {\bf A}_{\mathcal{G}_j}, {\bf Y}_j)$ from a training mini-batch ;
            \STATE Sample a data augmentation weight: $\beta \gets {\rm Beta}(\psi,\psi)$;
            \STATE Generate the augmented feature matrix $\hat{\bf X}_{\mathcal{G}_i\mathcal{G}_j} \gets \beta {\bf X}_{\mathcal{G}_i} + (1-\beta){\bf X}_{\mathcal{G}_j}$ and augmented adjacency matrix $\hat{\bf A}_{\mathcal{G}_i\mathcal{G}_j} \gets \beta {\bf A}_{\mathcal{G}_i} + (1-\beta){\bf A}_{\mathcal{G}_j}$;
            \STATE Generate the corresponding label: $\hat{\bf Y}_{ij} \gets \beta {\bf Y}_i + (1-\beta){\bf Y}_j$;
            \STATE ${\bf H}^{(0)} = \hat{\bf X}_{\mathcal{G}_i\mathcal{G}_j}$;
            \FOR{$l=0 \to L-1$}
                \STATE Sample a mask ${\bf Z}^{(l)}$ from Eq. \ref{Eq11};
                \STATE Generate the uncertain adjacency matrix: ${\bf A} \odot {\bf Z}^{(l)}$;
                \STATE Aggregate on each layer via Eq. \ref{Eq15};
            \ENDFOR
            \STATE Predict via Eq. \ref{Eq13} and calculate the total loss;
            \STATE Update the parameters by gradients descending;
        \ENDFOR
    \end{algorithmic}
\end{algorithm}

\section{Experiments and Results Analysis}
In this section, we provide details about the datasets used in our experiments and describe the experimental settings, as well as comparisons with state-of-the-art.

\subsection{Experimental Setting}
\subsubsection{Datasets}
Our investigation enlisted two established EEG datasets, the SJTU emotion EEG dataset (SEED) \cite{zheng2015investigating} and its successor SEEDIV dataset \cite{zheng2018emotionmeter}, in the execution of emotion recognition tasks. These datasets encompass EEG recordings from 15 youthful participants, evenly distributed by gender. The data, collected via a 62-channel ESI NeuroScan System as the subjects viewed emotionally charged film clips, were gathered over three sessions held on different days under uniform conditions to ensure data integrity. The raw EEG signals were then preprocessed, down-sampled from 1000 Hz to 200 Hz, and cleansed of artifacts stemming from eye movements (EOG) and muscle activity (EMG).

The SEED dataset comprises 15 film clips of around 4 minutes each, carefully crafted by psychologists to evoke specific emotional states - Positive, Neutral, or Negative. Hence, each participant in the SEED dataset produced 45 distinct EEG signal trials. The SEEDIV dataset, following a similar data acquisition procedure, increased the number of trials by featuring 24 film clips that lasted roughly 2 minutes each and elicited four types of emotions: Neutral, Sad, Fear, and Happy. This resulted in each participant contributing 72 EEG recording trials in the SEEDIV dataset.

\begin{table*}[t]
\renewcommand{\arraystretch}{1}
\setlength\extrarowheight{0.5 pt}
\caption{Mean Accuracy Rates (\%) and Standard Deviation of Subject-Dependent Test for the Four Kinds of Handcrafted Features Obtained From the Separate and Total Frequency Bands on SEED}
\label{tbl1}
\centering
\resizebox{\linewidth}{!}{\begin{tabular}{c|c|c|c|c|c|c|c}
\hline \hline
Feature & Method & $\delta$ band & $\theta$ band & $\alpha$ band & $\beta$ band & $\gamma$ band & All bands \\
\hline
\multirow{7}{*}{DE} & SVM \cite{zheng2015investigating}   & 60.50 / 14.14 & 60.95 / 10.20 & 66.64 / 14.41 & 80.76 / 11.56 & 79.56 / 11.38                     & 83.99 / 09.72 \\
                    & DBN \cite{zheng2015investigating}  & 64.32 / 12.45 & 60.77 / 10.42 & 64.01 / 15.97 & 78.92 / 12.48 & 79.19 / 14.58 & 86.08 / 08.34 \\
                    \cline{2-8}
                    & Bi-DANN \cite{li2018bi} & 76.97 / 10.95 & 75.56 / 07.88 & 81.03 / 11.74 & 89.65 / 09.59 & 88.64 / 09.46 & 92.38 / 07.04 \\
                    & R2G-STNN \cite{8736804} & 77.76 / 09.92 & 76.17 / 07.43 & 82.30 / 10.21 & 88.35 / 10.52 & 88.90 / 09.57 & 93.38 / 05.96 \\
                    & BiHDM \cite{li2020novel} & - & - & - & - & - & 85.40 / 07.53 \\
                    \cline{2-8}
                    & vGCN \cite{defferrard2016convolutional} & 72.75 / 10.85 & 74.40 / 08.23 & 73.46 / 12.17 & 83.24 / 09.93 & 83.36 / 09.43 & 87.40 / 09.20 \\
                    & DGCNN \cite{song2018eeg} & 74.25 / 11.42 & 71.52 / 05.99 & 74.43 / 12.16 & 83.65 / 10.17 & 85.73 / 10.64 & 90.40 / 08.49 \\
                    & RGNN \cite{zhong2020eeg} & 76.17 / 07.91 & 72.26 / 07.25 & 75.33 / 08.85 & 84.25 / 12.54 & 89.23 / 08.90 & 94.24 / 05.95 \\
                    & GCB-net+BLS \cite{zhang2019gcb} & 79.98 / 08.93 & 76.51 / 09.56 & 81.97 / 11.05 & 89.06 / 08.69 & 89.10 / 09.55 & 94.24 / 06.70 \\
                    & V-IAG \cite{song2021variational} & 81.14 / 09.46 & \textbf{82.37} / \textbf{07.44} & 84.51 / 09.68 & 92.15 / 08.90 & 92.96 / 06.19 & 95.64 / 05.08 \\
                    \cline{2-8}
                    & CU-GCN  & \textbf{82.01} / \textbf{08.76} & 80.62 / 08.33 & \textbf{84.92} / \textbf{10.32} & \textbf{92.56} / \textbf{09.17}  & \textbf{93.32} / \textbf{05.94} & \textbf{95.70} / \textbf{05.32} \\
\hline \hline
\multirow{7}{*}{PSD} & SVM \cite{zheng2015investigating} & 58.03 / 15.39 & 57.26 / 15.09 & 59.04 / 15.75 & 73.34 / 15.20 & 71.24 / 16.38 &                     59.60 / 15.93 \\
                    & DBN \cite{zheng2015investigating}  & 60.05 / 16.66 & 55.03 / 13.88 & 52.79 / 15.38 & 60.68 / 21.31 & 63.42 / 19.66 & 61.90 / 16.65 \\
                    \cline{2-8}
                    & vGCN \cite{defferrard2016convolutional} & 69.89 / 13.83 & 70.92 / 09.18 & 73.18 / 12.74 & 76.21 / 10.76 & 76.15 / 10.09 & 81.31 / 11.26 \\
                    & DGCNN \cite{song2018eeg} & 71.23 / 11.42 & 71.20 / 08.99 & 73.45 / 12.25 & 77.45 / 10.81 & 76.60 / 11.83 & 81.73 / 09.94 \\
                    & GCB-net+BLS \cite{zhang2019gcb} & 72.90 / 13.19 & 74.48 / 09.03 & 76.99 / 10.36 & 83.30 / 10.73 & 83.12 / 11.95 & 84.32 / 10.61 \\
                    & V-IAG \cite{song2021variational} & - & - & - & - & - & 86.71 / 10.25 \\
                    \cline{2-8}
                    & CU-GCN    & \textbf{74.10} / \textbf{09.78} & \textbf{73.84} / \textbf{07.35} & \textbf{77.00} / \textbf{09.53} & \textbf{84.56} / \textbf{10.49} & \textbf{85.49} / \textbf{09.17}  &  \textbf{87.21} / \textbf{09.81} \\
\hline \hline
\multirow{7}{*}{DASM} & SVM \cite{zheng2015investigating} & 48.87 / 10.49 & 53.02 / 12.76 & 59.81 / 14.67 & 75.03 / 15.72 & 73.59 / 16.57                      & 72.81 / 16.57 \\
                    & DBN \cite{zheng2015investigating}  & 48.79 / 09.62 & 51.59 / 13.98 & 54.03 / 17.05 & 69.51 / 15.22 & 70.06 / 18.14 & 72.73 / 15.93 \\
                    \cline{2-8}
                    & vGCN \cite{defferrard2016convolutional} & 57.07 / 06.75 & 54.80 / 09.09 & 62.97 / 13.43 & 74.97 / 13.40 & 73.28 / 13.67 & 76.00 / 13.32 \\
                    & DGCNN \cite{song2018eeg} & 55.93 / 09.14 & 56.12 / 07.86 & 64.27 / 12.72 & 73.61 / 14.35 & 73.50 / 16.60 & 78.45 / 11.84 \\
                    & GCB-net+BLS \cite{zhang2019gcb} & 62.36 / 10.66 & 65.00 / 10.31 & 70.91 / 10.84 & 85.55 / 11.39 & 86.04 / 10.85 & 82.09 / 13.14 \\
                    & V-IAG \cite{song2021variational} & - & - & - & - & - & 90.10 / 08.73 \\
                    \cline{2-8}
                    & CU-GCN    & \textbf{64.56} / \textbf{08.81} & \textbf{68.77} / \textbf{10.34} & \textbf{73.11} / \textbf{09.62} & \textbf{87.77} / \textbf{10.69} & \textbf{87.43} / \textbf{10.09}  &  \textbf{91.04} / \textbf{09.16} \\
\hline \hline
\multirow{7}{*}{RASM} & SVM \cite{zheng2015investigating}   & 47.75 / 10.59 & 51.40 / 12.53 & 60.71 / 14.57 & 74.59 /16.18 & 74.61 / 15.57                     & 74.74 / 14.79 \\
                    & DBN \cite{zheng2015investigating}  & 48.05 / 10.37 & 50.62 / 14.02 & 56.15 / 15.28 & 70.31 / 15.62 & 68.22 / 18.09 & 71.30 / 16.16 \\
                    \cline{2-8}
                    & vGCN \cite{defferrard2016convolutional} & 59.70 / 05.65 & 55.91 / 08.82 & 59.97 / 14.27 & 79.45 / 13:32 & 79.73 / 13.22 & 84.06 / 12.86 \\
                    & DGCNN \cite{song2018eeg} & 57.79 / 06.90 & 55.79 / 08.10 & 61.58 / 12.63 & 75.79 / 13.07 & 82.32 / 11.54 & 85.00 / 12.47 \\
                    & GCB-net+BLS \cite{zhang2019gcb} & 62.56 / 08.83 & 62.22 / 11.12 & 71.43 / 10.83 & 87.03 / 11.16 & 85.59 / 11.18 & 87.73 / 10.19 \\
                    & V-IAG \cite{song2021variational} & - & - & - & - & - & \textbf{90.53} / \textbf{09.22} \\
                    \cline{2-8}
                    & CU-GCN    & \textbf{63.98} / \textbf{07.21} & \textbf{64.52} / \textbf{08.79} & \textbf{72.70} / \textbf{11.32} & \textbf{88.82} / \textbf{09.92} & \textbf{86.56} / \textbf{10.72}  &  89.77 / 09.68 \\
\hline \hline
\end{tabular}}
\end{table*}

\subsubsection{Data Splits}

In the pursuit of equitable comparisons, we conducted both subject-dependent and subject-independent classifications on the SEED and SEED-IV datasets, in adherence with protocols laid out in previous investigations \cite{wang2014emotional, zheng2015investigating, song2018eeg, zhong2020eeg, song2021variational, jia2020sst}. 
In the subject-dependent trials on the SEED dataset, we designated the initial 3 trials from each emotional category (making up a total of 9 trials) for training purposes, while the residual 6 trials were assigned to the testing set, thereby ensuring an even distribution across categories. Notably, these experiments drew upon two sessions from each subject. We reported the final outcome as the mean accuracy across all 15 subjects.

Turning to the subject-dependent trials on the SEED-IV dataset, we allocated the first 4 trials from each emotional category (a total of 16 trials) for training, whereas the remaining 8 trials served as the testing set. All three sessions were incorporated into the evaluation of model performance. As with the SEED dataset, we reported accuracy over all 15 subjects.
Regarding the subject-independent trials, we espoused the leave-one-out methodology. Specifically, cross-validation was performed by excluding one session from each subject in the case of the SEED dataset, and all sessions from each subject in the case of the SEED-IV dataset. We evaluated model performance across all 15 trials.

\begin{table*}[t]
\renewcommand{\arraystretch}{1}
\caption{Mean Accuracy Rates (\%) and Standard Deviation of Subject-Dependent Test for the DE Feature Obtained From the Separate and Total Frequency Bands on SEEDIV}
\label{tbl2}
\centering
\resizebox{\linewidth}{!}{\begin{tabular}{c|c|c|c|c|c|c|c}
\hline \hline
Feature & Method & $\delta$ band & $\theta$ band & $\alpha$ band & $\beta$ band & $\gamma$ band & All bands \\
\hline
\multirow{7}{*}{DE} & SVM \cite{zheng2018emotionmeter}   & 57.58 / 12.64 & 57.98 / 12.30 & 61.22 / 16.46 & 66.66 / 18.80 & 66.34 / 17.49 &                     70.58 / 17.01 \\
                    & DBN \cite{zheng2015investigating}  & - & - & - & - & - & 66.77 / 07.38 \\
                    & BiDANN \cite{li2018bi} & - & - & - & - & - & 70.29 / 12.63 \\
                    & BiHDM \cite{li2020novel} & - & - & - & - & - & 74.35 / 14.09 \\
                    & vGCN$^{\dagger}$ \cite{defferrard2016convolutional} & 60.23 / 12.47  & 58.07 / 12.21  & 61.97 / 10.06  & 68.87 / 09.12  & 66.50 / 14.01  & 70.81 / 11.56 \\
                    & DGCNN$^{\dagger}$ \cite{song2018eeg} & 61.84 / 11.62  & 63.44 / 13.91  & 63.74 / 13.15  & 67.28 / 12.82  & 64.77 / 12.38  & 70.47 / 15.29  \\
                    & RGNN$^{\dagger}$ \cite{zhong2020eeg} & 64.23 / 09.83  & 62.33 / 10.14  & \textbf{65.98} / \textbf{08.52}  & 72.47 / 09.35  & 71.66 / 09.51  & 75.90 / 12.11   \\
                    \cline{2-8}
                    & CU-GCN & \textbf{66.21} / \textbf{07.71}  & \textbf{64.81} / \textbf{08.12} & 65.92 / 09.69 & \textbf{75.93} / \textbf{08.77} & \textbf{73.44} / \textbf{10.10} & \textbf{77.39} / \textbf{09.24} \\
\hline \hline
\end{tabular}}
\begin{tablenotes}
    \footnotesize
    \item[1] $\dagger$ indicates  the  experiment  results  obtained  are  based  on  our  own implementation.
\end{tablenotes}
\end{table*}

\begin{table*}[t]
\renewcommand{\arraystretch}{1}
\setlength\extrarowheight{0.5 pt}
\caption{Mean Accuracy Rates (\%) and Standard Deviation of Subject-Independent Test with the DE Feature Obtained From the Separate and Total Frequency Bands on SEED and SEEDIV}
\label{tbl3}
\centering
\resizebox{\linewidth}{!}{\begin{tabular}{c|c|c|c|c|c|c|c}
\hline \hline
    \multirow{2}{*}{Method} & \multicolumn{6}{c}{SEED} & SEEDIV \\
    \cline{2-8}
    & $\delta$ band & $\theta$ band & $\alpha$ band & $\beta$ band & $\gamma$ band & All bands & All bands \\
    \hline
    SVM \cite{zheng2015investigating}   & 43.06 / 08.27 & 40.07 / 06.50 & 43.97 / 10.89 & 48.63 / 10.29 & 51.59 / 11.83 & 56.73 / 16.29 & 37.99 / 12.52 \\
    BiDANN-S \cite{li2018bi} & 63.01 / 07.49 & 63.22 / 07.52 & 63.50 / 09.50 & 73.59 / 09.12 & 73.72 / 08.67 & 84.14 / 06.87 & 65.59 / 10.39 \\
    DANN \cite{ganin2016domain} & 56.66 / 06.48 & 54.95 / 10.45 & 59.37 / 10.57 & 67.14 / 07.10 & 71.30 / 10.84 & 75.08 / 11.18 & - \\
    R2G-STNN \cite{8736804} & 63.34 / 05.31 & 63.78 / 07.53 & 64.27 / 10.88 & 74.85 / 08.02 & 74.54 / 08.41 & 84.16 / 07.63 & - \\
    BiHDM \cite{li2020novel} & - & - & - & - & - & 85.40 / 07.53 & 69.03 / 08.66 \\
    DGCNN \cite{song2018eeg} & 49.79 / 10.94 & 46.36 / 12.06 & 48.29 / 12.28 & 56.15 / 14.01 & 54.87 / 17.53 & 79.95 / 09.02 & 52.82 / 09.23 \\
    RGNN \cite{zhong2020eeg} & 64.88 / 06.87 & 60.69 / 05.79 & 60.84 / 07.57 & 74.96 / 08.94 & 77.50 / 08.10 & 85.30 / 06.72 & 73.84 / 08.02 \\
    V-IAG \cite{song2021variational} & - & - & - & - & - & \textbf{88.38} / \textbf{04.80} & - \\
    \hline
    CU-GCN    & \textbf{64.72} / \textbf{05.99} & \textbf{64.04} / \textbf{07.57} & \textbf{64.50} / \textbf{09.46} & \textbf{76.92} / \textbf{07.76} & \textbf{79.38} / \textbf{09.03} & 87.10 / 05.44 & \textbf{74.50} / \textbf{07.88} \\
\hline \hline
\end{tabular}}
\begin{tablenotes}
    \footnotesize
    \item[1] $\dagger$ indicates the experiment results obtained are based on our own implementation.
\end{tablenotes}
\end{table*}
\subsubsection{Model Training}
We conducted our experiments using the PyTorch Geometric toolbox and trained the models on a single NVIDIA Tesla A100 GPU, utilizing the Adam optimizer. To determine the optimal hyperparameters, we performed a search on the validation set, exploring the following ranges: (a) initial learning rate within the range [1e-3, 3e-2]; (b) the number of convolutional layers within range \{2, 4, 6, 8\} and hidden feature dimension within range \{32, 64, 128\}. We set the temperature parameter $t$ in the concrete distribution as 0.67, the Beta distribution weight $\psi$ as 2, and the batch size as 32. The hyper-parameters were hand-tuned with the best performance over the validation set. A cosine annealing learning rate scheduler was adopted for all experiments. We adopt an early stopping strategy when the validation loss did not decrease for ten consecutive epochs. 

\subsection{Experimental Results}
%
We report the mean accuracy (Acc) and standard deviation (Std) as the main evaluation metrics for emotion recognition. This choice aligns with recent studies in the field \cite{wang2014emotional, zheng2015investigating, song2018eeg, zhong2020eeg, song2021variational, jia2020sst}, ensuring consistency in the evaluation of our results.

\subsubsection{Subject-dependent Test}
Following the procedure of Algorithm \ref{alg1}, we compare the proposed model with a quantity of popular used models: (1) traditional machine learning classifying models, e.g. SVM \cite{zheng2018emotionmeter}, DBN \cite{zheng2015investigating}; (2) non-topological deep learning methods, e.g. CNN alone or combined with RNN, such as BiDANN \cite{li2018bi}, BiHDM \cite{li2020novel}  and R2G-STNN \cite{8736804}; (3) recently proposed GNN models: vanilla GCN (vGCN) \cite{kipf2016semi}, DGCNN \cite{song2018eeg}, RGNN \cite{zhong2020eeg}, GCB-net with BLS \cite{zhang2019gcb} and V-IAG \cite{song2021variational}. To further explore the relation of spectral information and emotion state, we conduct these experiments from the separate and total five frequency bands, known as $\delta$ band (1-4 Hz), $\theta$ band (4-8 Hz), $\alpha$ band (8-14 Hz), $\beta$ band (14-30 Hz) and $\gamma$ band (30-50 Hz). For a fair comparison, we adopt the pre-extracted DE, PSD, DASM and RASM features after linear dynamic systems (LDS) as our input source data. 

Table \ref{tbl1} shows the performance of our model and the SOTA methods. In view of the DE feature is more effective in emotion recognition, we only take experiments with DE feature on SEEDIV, with results shown in Table \ref{tbl2}. Our model performs on par or better than other methods both on separate and total frequency bands on all features. We analyze the results from three aspects: (1) input source - DE feature are experimentally proved as the most effective metric; (2) classifier - deep learning methods among which CNN is used to acquire spatial attention and RNN is used to learn temporal relation, outperformed traditional machine learning methods for a lot. While GCN-based methods utilize the EEG topological structure and make a more reasonable channel grouping, finally present the best performance; (3) spectral effectiveness: leverage features on $\beta$ and $\gamma$ band alone are more effective for emotion activation but greatly under-performed than using features on whole bands, somehow indicating that high-frequency information is more active with emotional activities. Our model obtained the best performance on both datasets. We attribute such a performance gain to the leverage of uncertainty-guided instance-wise adjacency matrix and the improvement of model ability on the spectral richness and alleviation to over-smoothing.

\subsubsection{Subject-independent Test}
Physiological signals, such as EEG, present severe individual variation and greatly hinder the model generalization. Thus, we conduct the subject-independent emotion classification to further verify our model stability. We use DE feature as the exclusive input for its remarkable discriminating property. Table \ref{tbl3} shows the performance of three categorical classifiers on SEED and SEEDIV. Note that for brief comparison, we conduct the test using features from all five bands on SEEDIV following previous literature settings. Not surprisingly, despite numerical value presenting a similar tendency as subject-dependent experiments in the view of all three analysis aspects, all accuracy suffered a great decrease. Our method achieves the best performance in SEEDIV and comparable results in the SEED datasets, which verifies the generalization of CU-GCN with subject-independent EEG emotion recognition. R2G-STNN leveraged a bidirectional long short term memory (BiLSTM) to learn the region to global spatial-temporal information. BiHDM proposed a four-directed RNN based on two spatial orientations to obtain the discrepancy information between two separate brain regions and used a domain discriminator to generate the domain-invariant feature. RGNN eliminates the individual invariant by introducing a similar domain adversarial training pattern. V-IAG was believed to promote cross-subject performance by designing a more reasonable adjacency matrix that contains an instance adaptive branch and a variational branch to learn the deterministic graphs and an uncertain graph. All these methods solve the subject-independent problem by designing a data-driven node connection graph from both regional and global space. It indicates that the key point of emotion recognition should lie in the precise combination of EEG channels.

\begin{figure}[t]
	\centering
	\subfigure[]{
	\begin{minipage}[t]{\linewidth}
	    \centering
	    \centerline{\includegraphics[width=2.7in]{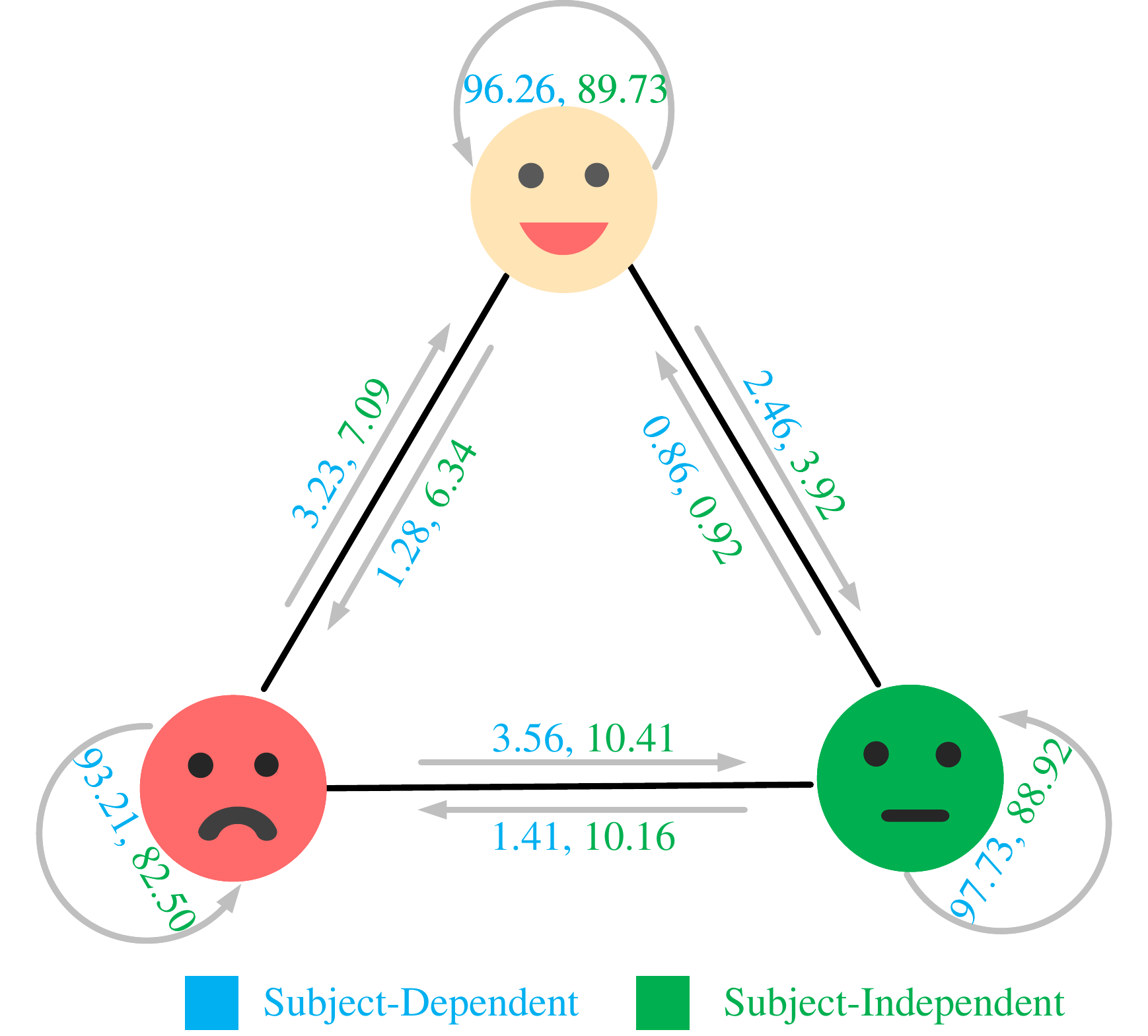}}
	    \label{FIG:3a}
	\end{minipage}}
 	\subfigure[]{
	\begin{minipage}[t]{\linewidth}
	    \centering
	    \centerline{\includegraphics[width=2.7in]{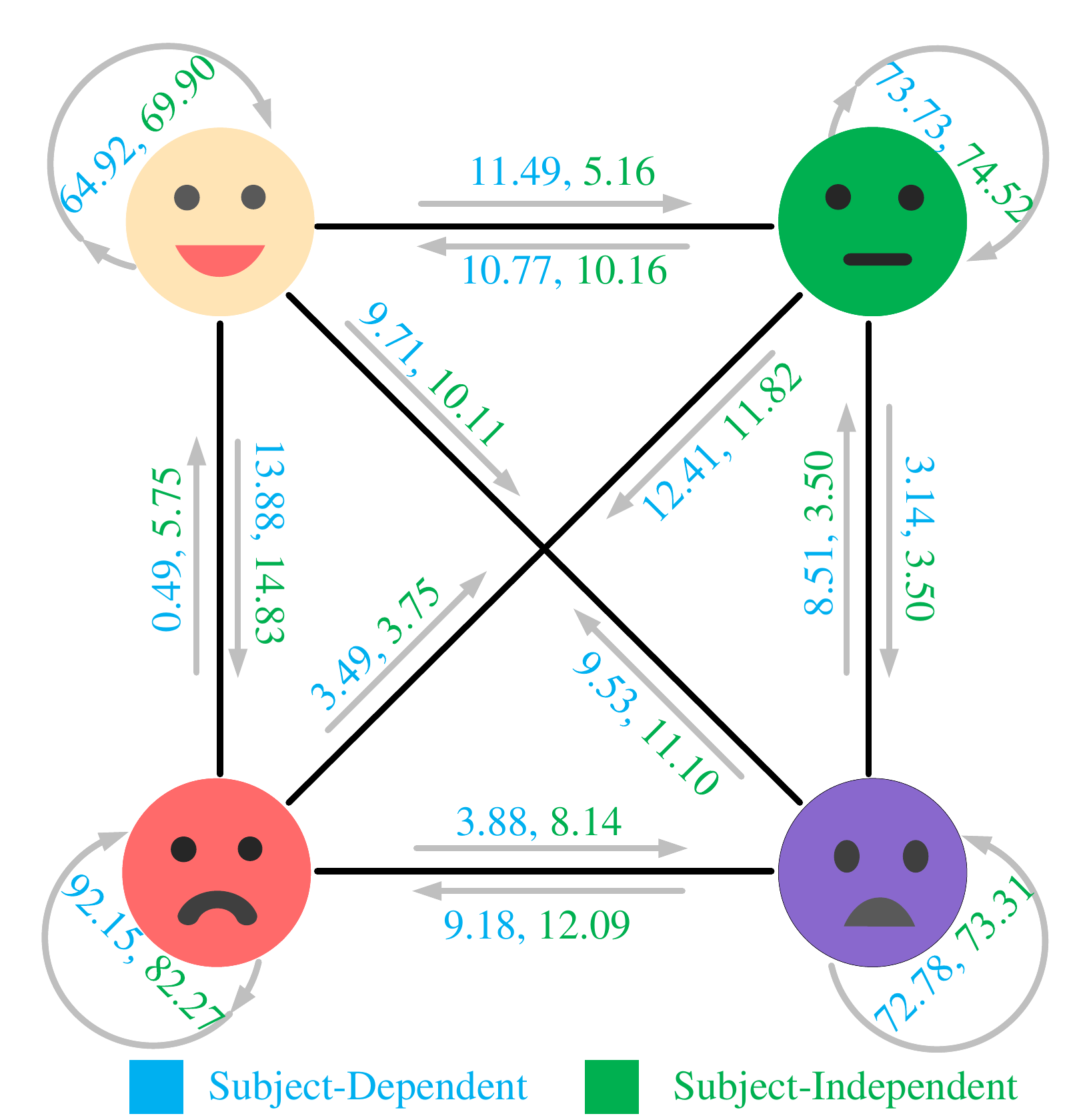}}
	    \label{FIG:3b}
	\end{minipage}
	}
\centering
\caption{Confusion Graph of Subject-Dependent/Subject-Independent Experiments on SEED and SEEDIV Datasets.}
\label{FIG:3}
\end{figure}

\begin{table}[t]
\renewcommand{\arraystretch}{1}
\caption{Ablation Study Results for Subject-Independent Emotion Recognition on SEED and SEEDIV Datasets. Mean accuracy (\%) and standard deviations are reported.}
\label{tbl4}
\centering
\resizebox{\linewidth}{!}{\begin{tabular}{l|c|c}
\hline \hline
Method & SEED & SEEDIV \\
\hline \hline
proposed & 87.10 / 05.44 & 74.50 / 07.88\\
\hline \hline
    \quad - eGCN $\rightarrow$ GCN(2)  & 83.82 / 07.33  & 72.13 / 09.41 \\
    \quad - eGCN $\rightarrow$ SGC(2)  & 83.43 / 07.18  & 72.59 / 10.26 \\
\hline \hline
    \quad - Dist adjacency matrix & 85.66 / 06.73 & 71.37 / 09.01 \\
    \quad - Coh adjacency matrix & 85.10 / 07.13 & 72.35 / 08.20 \\
    \quad - Random adjacency matrix & 85.29 / 07.03 & 71.93 / 08.77 \\
\hline \hline
    \quad - w/o Data-aug  & 85.47 / 05.82  & 72.61 / 08.50\\
\hline \hline
\end{tabular}}
\end{table}

\begin{table}[t]
\renewcommand{\arraystretch}{1}
\setlength\extrarowheight{1 pt}
\caption{Comparison of over-Smoothing elimination ability between CU-GCN (w/o EP) and existing classic GCNs in subject-independent emotion recognition on SEED dataset.}
\label{tbl5}
\centering
\setlength{\tabcolsep}{1mm}{
\begin{tabular}{c|c|c|c|c}
\hline \hline
 \multirow{2}{*}{Method} & \multicolumn{4}{c}{Layers ($ L$)} \\
\cline{2-5}
       & 2 & 4 & 6 & 8  \\
\hline
    vGCN & 77.85 / 10.18 & 76.12 / 10.33 & 71.00 / 12.53 & 65.96 / 09.51 \\
    SGC & 76.11 / 11.92 & 75.48 / 11.72 & 70.32 / 11.02 & 63.58 / 15.58 \\
    CU-GCN & 84.12 / 05.72 & 84.58 / 05.33 & 85.71 / 05.53 & 84.66 / 05.98 \\
\hline \hline
\end{tabular}
}
\end{table}

\subsection{Ablation Studies and Discussions}
We conducted ablation studies to analyze and discuss the components of our CU-GCN model and their impact on subject-independent emotion recognition.

\subsubsection{Graph Neural Network Performance}
GCN-based methods from Table \ref{tbl1}, \ref{tbl2}, \ref{tbl3}, including DGCNN, RGNN, and V-IAG, have demonstrated superior performance in EEG-based emotion recognition tasks compared to other approaches. DGCNN pioneered the utilization of two-layer vanilla GCN to process EEG signals by leveraging graph structure. Building upon this, RGNN further emphasized the importance of node aggregation in graph structure learning and employed a two-layer Spectral Graph Convolution (SGC) as the backbone. However, both vanilla GCN and SGC suffer from the over-smoothing problem, which arises as the network depth increases, leading to the loss of local information. V-IAG employed an eight-layer GCN architecture that preserved features from local to global regions by aggregating the outputs of each layer for the final decision.

In light of our previous theoretical analysis, our enhanced GCN (eGCN) is designed to address the over-smoothing problem while enhancing spectral information. In this subsection, we perform ablation studies by replacing eGCN with SGC and GCN to validate our hypothesis. Table \ref{tbl4} presents the results, where "eGCN $\rightarrow$ GCN(2)" indicates the replacement of eGCN with a two-layer GCN, and similarly for SGC. As shown in Table \ref{tbl4}, it is evident that eGCN achieves significantly better results than the common benchmark methods.

Furthermore, we conduct additional experiments by varying the number of layers ($L$ corresponding to the number of aggregated filters in CU-GCN). The results in Table \ref{tbl5} consistently demonstrate that our method outperforms others, with the best performance achieved when $L=6$, equivalent to six layers. These findings indicate that eGCN effectively mitigates the over-smoothing problem as the network depth increases. Notably, the aggregation of information from six-hops neighborhoods performs optimally, potentially due to ensuring the global connection between two hemisphere brain regions, as the relative node distance in a 2D squared matrix is nine. Overall, these results highlight the superior ability of CU-GCN to aggregate information over larger neighborhoods compared to other methods.

\subsubsection{Comparison Between Graph Structures}
To compare the effectiveness of our uncertainty-guided edge predictor (EP) with commonly used graph construction approaches, we trained CU-GCN for subject-independent emotion recognition on the SEED and SEEDIV datasets. The second part of Table 5 presents the performance of CU-GCN with and without EPM using different adjacency matrices: distance-based (Dist), coherence-based (Coh), and random adjacency matrices. We observed that the inclusion of EP consistently yielded better results compared to other methods. Interestingly, traditional graph construction methods did not exhibit significant differences, suggesting that EEG signals in emotion recognition possess unique characteristics. Our method offers several distinct advantages:
(a) It can be applied even when the physical locations of electrodes are unknown, providing flexibility in practical scenarios.
(b) It captures dynamic brain connectivity patterns instead of relying solely on spatial sensor information, which is particularly desirable for emotion recognition tasks.
(c) It enables adaptive selection of spectral features in EEG, facilitating precise categorization.

\subsubsection{What Is The Effect of Concrete Bernoulli Distribution?}

In our study, we conducted an investigation into the impact of using a learnable Bernoulli prior distribution compared to a constant edge dropout method. The learnable Bernoulli success rate $p$ in our CU-GCN model is adaptively optimized during training. The results presented in Table \ref{tbl6} provide compelling evidence for the effectiveness and necessity of this approach, as CU-GCN with the learnable Bernoulli prior consistently outperforms the constant $p$ value method in terms of classification accuracy.
Moreover, it is worth noting that even without the learnable Bernoulli prior, our CU-GCN approach still achieves competitive performance, often surpassing the results obtained by state-of-the-art methods. This highlights the robustness and effectiveness of CU-GCN in capturing important graph structure information and achieving accurate emotion classification, even in the absence of the learnable prior distribution.
Overall, these findings emphasize the significance of incorporating the learnable Bernoulli prior distribution in CU-GCN, as it contributes to improved classification accuracy and enhances the model's ability to leverage graph structure for emotion recognition tasks.

\begin{table}[t]
\renewcommand{\arraystretch}{1}
\setlength\extrarowheight{1 pt}
\caption{Adaptive vs. Predefined Bernoulli Probability: Performance Comparison on Subject-Independent Emotion Recognition.}
\label{tbl6}
\centering
\setlength{\tabcolsep}{3mm}{
\begin{tabular}{c|c|c|c|c|c}
\hline \hline
\cline{1-6}
      p & 0.3  & 0.5 & 0.7 & 0.9 & Adaptive  \\
\hline
Accuracy (\%) & 85.11 & 85.73  & 85.92 & 85.64 & 87.10\\
\hline \hline
\end{tabular}
}
\end{table}

\subsubsection{Is the Graph Data mixup Method Effective?}
In our CU-GCN framework, we employ the graph mixup data augmentation technique to enhance the probability confidence and facilitate the learning process in the presence of noisy labels. We conducted experiments to evaluate the effectiveness of this approach, and the results are presented in Table \ref{tbl4}. The findings indicate that training with graph mixup leads to an improvement of 0.2$ \sim $0.3\% in performance compared to training with the original dataset alone. Furthermore, the confusion matrices depicted in Figure \ref{FIG:3} for the SEED and SEEDIV datasets demonstrate the substantial enhancement achieved by our data augmentation methods in addressing the challenges posed by noisy labels.

\section{Conclusion}

In summary, our proposed method, which combines uncertainty-guided graph modeling and spectral-enhanced message aggregation, along with a noisy label learning approach, has achieved significant advancements in EEG-based emotion recognition. We have established new state-of-the-art performance in both subject-dependent and subject-independent emotion classification tasks using two publicly available datasets. Our method demonstrates improved robustness and generalization capabilities. Notably, we have observed that instance adaptive graph learning methods yield more accurate recognition of emotional states compared to simply considering channel mutual relationships. Moreover, our extensive visualization experiments provide valuable support for distributionist theories in brain neuroscience. Looking ahead, our study opens up exciting possibilities for applying graph-based representations in multi-modal emotion recognition applications, as our methods are not limited to EEG alone.


\section*{Acknowledgments}
This research was funded by the National Natural Science Foundation of China (62171123, 62001105, 62211530112, and 62071241), the China Postdoctoral Science Foundation (2023M730585), the National Key Research and Development Program of China (2022YFC2405600), the Natural Science Foundation of Jiangsu Province (BK20192004), the Jiangsu Funding Program for Excellent Postdoctoral Talent (2023ZB812) and the Postgraduate Research \& Practice Innovation Program of Jiangsu Province (KYCX20\underline{ }0088).

\section*{References}
\bibliographystyle{IEEEtran}
\bibliography{ref.bib}

\end{document}